\documentclass[runningheads]{llncs}
\usepackage{epsfig}
\usepackage{graphicx}
\usepackage{amsmath}
\usepackage{amssymb}
\usepackage{color}
\usepackage{mathrsfs}  
\usepackage{amssymb,amsmath,bm}
\usepackage{multicol,multicap,graphicx}
\usepackage{subcaption}
\usepackage{wrapfig}
\usepackage{picinpar}
\usepackage{cutwin}
\usepackage{algorithm}  
\usepackage{algorithmic}
\usepackage{amsfonts}
\usepackage{amsfonts,amssymb}
\usepackage{url}
\usepackage{hyperref}
\usepackage{stmaryrd}

\DeclareMathOperator*{\argmin}{arg\,min}
\usepackage{mathtools}
\usepackage{enumitem}
\begin{document}
\title{Adversarial Regression Learning \\for Bone Age Estimation}
%
%\titlerunning{Abbreviated paper title}
% If the paper title is too long for the running head, you can set
% an abbreviated paper title here
%
% \author{Anonymous}
% %\institute{Anonymous Address}
% \institute{}
\author{Youshan Zhang  \ \and   \ Brian D. Davison}
% \institute{
% University of Jane, Italy\\
% \url{jane@university.it}
% \and
% University of John, Germany\\
% \url{john@university.de} 
% }

\institute{Lehigh University, Computer Science and Engineering, Bethlehem, PA, USA\\
\email{\{yoz217,bdd3\}@lehigh.edu}}

% \author{First Author\inst{1}\orcidID{0000-1111-2222-3333} \and
% Second Author\inst{2,3}\orcidID{1111-2222-3333-4444} \and
% Third Author\inst{3}\orcidID{2222--3333-4444-5555}}
% %
% \authorrunning{F. Author et al.}
% % First names are abbreviated in the running head.
% % If there are more than two authors, 'et al.' is used.
% %
% \institute{Princeton University, Princeton NJ 08544, USA \and
% Springer Heidelberg, Tiergartenstr. 17, 69121 Heidelberg, Germany
% \email{lncs@springer.com}\\
% \url{http://www.springer.com/gp/computer-science/lncs} \and
% ABC Institute, Rupert-Karls-University Heidelberg, Heidelberg, Germany\\
% \email{\{abc,lncs\}@uni-heidelberg.de}}
%
\maketitle              % typeset the header of the contribution
\begin{abstract}
Estimation of bone age from hand radiographs is essential to determine skeletal age in diagnosing endocrine disorders and depicting the growth status of children. However, existing automatic methods only apply their models to test images without considering the discrepancy between training samples and test samples, which will lead to a lower generalization ability.
%to the test data.  
In this paper, we propose an adversarial regression learning network ($ARLNet$) for bone age estimation. Specifically, we first extract bone features from a fine-tuned Inception V3 neural network and propose regression percentage loss for training.  To reduce the discrepancy between training and test data, we then propose adversarial regression loss and feature reconstruction loss to guarantee the transition from training data to test data and vice versa, preserving invariant features from both training and test data.  Experimental results show that the proposed model outperforms state-of-the-art methods. 

\end{abstract}
\section{Introduction}

Skeletal age estimation from hand radiology images is extensively used in endocrinological disease diagnosis, judgment of children's growth, and genetic disorder diagnoses~\cite{escobar2019hand}. Bone age basically reflects the appearance of the hand.  As children grow, their bones become longer and change from cartilage to proper bones, such that we can estimate how old a child is 
%by looking at his bone images 
based on the average age of children with similar bone images.

% In normal development, children's bone age slightly fluctuates with their physical age (chronological age)~\cite{de2014hand}.  There are exceptions. 
% A serious mismatch between bone age and chronological age is a sign of physical problems, such as growth disorders and endocrine problems. Bone age can not only tell doctors and patients the relative maturity of the bone at a specific time but also combine with other clinical indicators to distinguish between normal and abnormal growth rates. Continuous bone age readings may indicate a child's developmental direction or progress during treatment. By assessing bone age, pediatricians can diagnose endocrine and metabolic disorders in children's development, such as bone dysplasia or growth defects affected by nutrition, metabolism, or other unknown factors, which may damage epiphyses or bone maturation. In the case of growth retardation, bone age and height may have the degree of delay, but after treatment, these children can still reach normal adult heights~\cite{creo2017bone,de2014hand}. Therefore, bone age estimation is important in evaluating the development and degree of ossification in the epiphysis.

Bone age assessment starts with taking  X-ray images of children's hands. For decades, the assessment of bone age was usually based on manual visual assessment of bone development of palms and wrists \cite{bayer1959radiographic,tanner2001assessment}.
%(Greulich and Pyle procedure~\cite{bayer1959radiographic} and  Tanner-Whitehouse~\cite{tanner2001assessment} technique). 
Hence, it is tedious and error prone. Therefore, it is necessary to develop an accurate automatic bone age algorithm. Recently, deep neural networks have become widely used in the medical field. Automatic bone age estimation using deep learning methods is much faster than manual labeling, while accuracy exceeds conventional methods~\cite{larson2018performance}. Most existing deep learning based methods take advantage of well-trained models on ImageNet datasets such as VGG16~\cite{simonyan2014very}, ResNet50~\cite{he2016deep}, or Inception V3~\cite{szegedy2016rethinking} for feature extraction, and add a regression layer to output bone age. 

Although many methods achieve effective results in bone age estimation, they still suffer from one challenge:  models are only optimized with training data, while the differences between training and test samples are omitted. Such models hence have a lower generalization ability to test samples if they are different from training data. To address this challenge, we aggregate three different loss functions in one framework:  regression percentage error loss,  adversarial regression loss, and feature reconstruction loss.

The contributions of this paper are three-fold:
\vspace{-0.1cm}
\begin{enumerate}
     \item To reduce the data discrepancy, we first extract features from a fine-tuned Inception V3 neural network  and propose a regression percentage error loss, which addresses both individual prediction error and mean prediction error; 
    \item To the best of our knowledge, we are the first to propose an adversarial regression loss to reduce the data difference during the training;
    \item The proposed feature reconstruction loss is able to maintain the feature consistency information of training and test samples.
\end{enumerate}
Extensive experiments are conducted on a large-scale Bone-Age dataset and achieve superior results over state-of-the-art methods. Our architecture is also successfully applied to another regression task: face age estimation.

\section{Related work}

%
% Manual assessment requires an expert's knowledge and considerable time.  BoneXpert~\cite{thodberg2008bonexpert} is a bone age assessment tool.  It is based on an active appearance model, which reconstructs the contours of 13 bones of a hand and determines bone age based on shape, texture, and intensity. However, this software expects good quality images (not always possible in practice, as shown in Fig.~\ref{fig:exm}) and not all bones are analyzed. 

In recent decades, bone age estimation has changed from manual assessment to automatic estimation algorithms. Deep learning methods have proved to be better than traditional machine learning methods in estimating in bone age. 

Halabi et al.~\cite{halabi2019rsna} reported the best five models for bone age estimation of the Radiological Society of North America (RSNA) challenge. The best model utilized the Inception V3 architecture concatenated with gender information. Data augmentation was leveraged to prevent overfitting and improve performance. The best performance achieves an error of 4.2 months (mean absolute difference, or MAD) in the challenge. 
% Other models uniformly extracted overlapping patches or first segmented the bones to localize the important regions of bones and then trained a regression  model; they obtained 4.35 and 4.50 months MAD, respectively~\cite{halabi2019rsna}.
% Larson et al.~\cite{larson2018performance} first processed images using contrast-limited adaptive histogram equalization and random augmentation. They fine-tuned a ResNet50 model (pre-trained on the ImageNet dataset) and their model achieved 6.0 months MAD.
% They found that the estimation of their model was close to expert assessment.
Iglovikov et al.~\cite{iglovikov2018paediatric} first used a dilated U-net to segment hand, and then they removed the background and normalized images. They finally calculated the affine transformations to register images. The bone age assessment model consists of a VGG-style regression model and classification model. The classification model aimed to output the bone age class, while the regression model output the bone age. 
% The bone age was reported based on the ensemble of three regions (whole hand, carpal bones, metacarpals and proximal phalanges).
The model achieved 4.20 months MAD. Chen et al.~\cite{chen2020attention}  proposed an attention-guided approach to automatically localize the discriminative regions to estimate bone age. 
% The authors first employed a classification model to find attention heat maps of discriminative regions (carpal bones and metacarpal bones). 
They then aggregated different regions for bone age assessment. Their model achieved 4.7 months MAD. Similarly, Escoba et al.~\cite{escobar2019hand} developed an approach to focus on local analysis of anatomical ROIs. They presented the additional bounding boxes and ROIs annotations during training and proposed a hand detection and hand pose estimation model to extract local information for bone age estimation. 
% Koitka et al.~\cite{koitka2020mimicking} built an automated system (including a detector network and a regression network) for pediatric bone age estimation. The detector network identified the ossification areas of bone. The regression network can output the bone age based on detected areas, and they obtained 4.56 months MAD. 
Pan et al.~\cite{pan2019improving} combined four different less-correlated and relatively high-performing models, and  they found that creating a ensemble model based on several weak models convincingly outperformed single-model prediction for bone age assessment. 

\begin{figure*}[t]
\centering
\begin{subfigure}{0.48\textwidth}
\includegraphics[width=\linewidth]{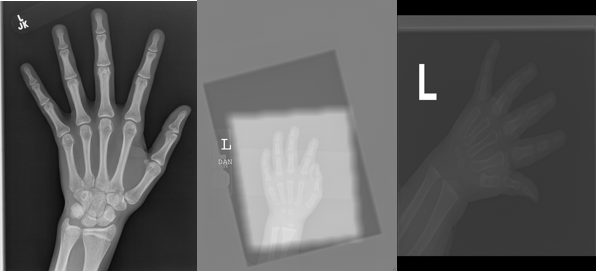}
\caption{Sample training images}
\end{subfigure}
\begin{subfigure}{0.49\textwidth}
\includegraphics[width=\linewidth]{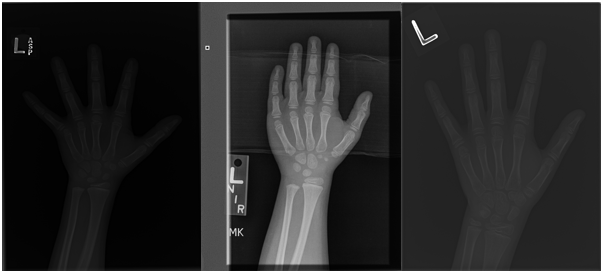}
\caption{Sample test images (can be quite faint)}
\end{subfigure}
\caption{Some bone radiology images from training and test data (differences between the training and test samples cause the difficulty of transfer learning).} \label{fig:exm}
\vspace{-0.4cm}
\end{figure*}

% \vspace{-0.1cm}
Transfer learning applies existing knowledge to a new domain. Although many methods take advantage of pre-trained ImageNet models to estimate the bone age, they do not consider the differences between training and testing data, which can sometimes be quite obvious, as shown in Fig.~\ref{fig:exm}. The discrepancy of data (named data shift/bias) will cause a vulnerability in the trained model, which causes poorer generalization on test data~\cite{zhang2020impact,zhang2020domain}. Transductive transfer learning uses both labeled training samples and unlabeled test samples to train the model and then uses a trained model to infer the label of the unlabeled test set. Hence, the data bias is mitigated via such a paradigm, which can improve the performance of the test data~\cite{zhang2019transductive}. In this paper, we propose adversarial regression learning to estimate bone age and simultaneously reduce the data shift between training and test datasets. 

\section{Method}
\subsection{Motivation}
As shown in Fig.~\ref{fig:exm}, we observe differences between training and test data. Unlike previous work that only optimizes the model based on training data, we utilize the idea of transductive learning---we train the neural network using both the labeled training and unlabeled test data, which reduces the discrepancy between them. We adopt learning theory from~\cite{ben2010theory} that the test risk can be minimized via bounding the training risk and discrepancy between them as follows.  
   
\textbf{Theorem 1}  Let $h$ be a hypothesis, $\epsilon_{tr} (h)$ and $\epsilon_{te} (h)$ represents the training and test risk (or error), respectively.
\begin{equation}
\begin{aligned}\label{eq:error}
\epsilon_{te} (h) \leq \epsilon_{tr} (h) + d_{\mathcal{H}} (\mathcal{D}_{tr}, \mathcal{D}_{te}) + C 
\end{aligned}
\end{equation}
where $d_{\mathcal{H}} (\mathcal{D}_{tr}, \mathcal{D}_{te})$ is the $\mathcal{H}$-divergence of training and test data, $C$ is the adaptability to quantify the error in an ideal hypothesis space of training and test data, which should be a sufficiently small constant. 

\subsection{Problem}
Bone age estimation is a regression problem. Given training data $\mathcal{X}_{tr}$ (including bone radiology image and gender information) with its labels $\mathcal{Y}_{tr} \in (0, \mathbb{R}^{+})^{N_{tr}}$ and test data $\mathcal{X}_{te}$ without its labels ($\mathcal{Y}_{te} \in (0, \mathbb{R}^{+})^{N_{te}}$ for evaluation only), the goal in bone age estimation is to learn a regressor $\mathcal{R}$  to minimize the test data risk and reduce the discrepancy between training and test data.

For most existing models, in the absence of data shift, regression models simply learn a regressor $\mathcal{R}$ that performs the task on training data, and minimizes the loss function in Eq.~\ref{eq:training}:   
\begin{equation}\label{eq:training}
\mathcal{L}_{tr} (\mathcal{X}_{tr}, \mathcal{Y}_{tr})= \mathbb{E} [ \ell ( \mathcal{R} (\mathcal{X}_{tr}),  \mathcal{Y}_{tr})  ],
\end{equation}
where $\mathbb{E} [\cdot]$ is the expectation and $\ell$ can be any appropriate loss function. Eq.~\ref{eq:training} only minimizes the training risk $\epsilon_{tr} (h)$. 

%In our model, training and test features are from feature extractor $G$. 
We define $\ell$  as the regression percentage error loss, $\ell =  \mathcal{L_P}= \mathcal{L_M} + \alpha\mathcal{L_D}$, where $\mathcal{L_M}$ is the mean absolute percentage error loss, $\mathcal{L_D}$ is the proposed absolute mean discrepancy loss and $\alpha$ is the balance factor between two loss functions. Specifically, $
 \mathcal{L_M} = \frac{1}{N_{tr}}\sum_{i=1}^{N_{tr}} |\frac{\mathcal{Y}_{tr_{i}} -\mathcal{Y}_{tr_{i}}'}{\mathcal{Y}_{tr_{i}}}|,  \ \ \ \mathcal{L_D}   =  |\frac{\overline{\mathcal{Y}_{tr}} -\overline{\mathcal{Y}_{tr}'}}{\overline{\mathcal{Y}_{tr}}}|
$,
where $\overline{\mathcal{Y}_{tr}}$ and $\overline{\mathcal{Y}_{tr}'}$ denote the mean actual and predicted bone age in training dataset. For training data, such a loss $\ell$ can capture both individual and mean percentage error.

In Eq.~\ref{eq:training}, the loss of training data did not reduce the data shift issue. Therefore, previous works have lower generalization to the test data. To mitigate the effects of data shift, we follow previous adversarial learning~\cite{ghifary2014domain,tzeng2017adversarial} to map samples across different domains to a common space,
% such that an adversarial discriminator is unable to distinguish the domains
and then invariant features are maintained. 
Hence, our model can learn on the training data, while retaining high generalization ability on test data.  

\begin{figure*}
    \centering
    \includegraphics[width=0.8\textwidth]{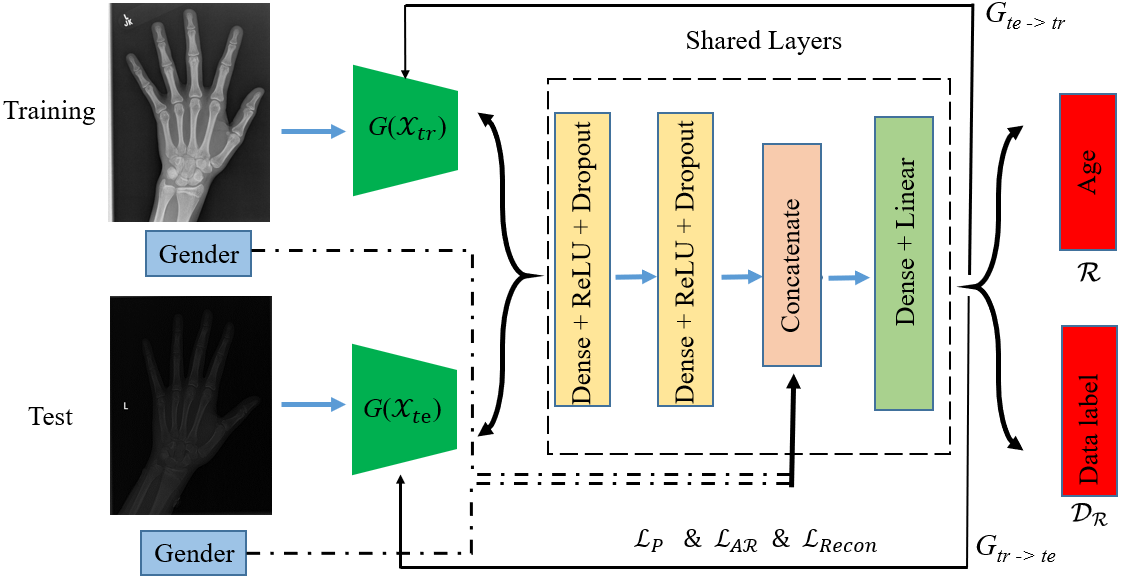}
    \caption{The architecture of our proposed adversarial regression learning network ($ARLNet$). Training and test features are first extracted via $G$. Then, $G(\mathcal{X}_{tr})$ and $G(\mathcal{X}_{te})$ are fed into shared layers.  The concatenate layer combines features from preceding layers and the gender. The weights of shared layers are updated by labeled training and unlabeled test data. $\Phi_{tr\shortrightarrow te}$ and $\Phi_{te\shortrightarrow tr}$ is the reconstruction from training data to test data and from test data to training data using hidden reconstruction layers (the detailed reconstruction layers are in Fig.~\ref{fig:model_more}). $\mathcal{L_P}$: regression percentage error loss, $\mathcal{L_{AR}}$: adversarial regression loss, $\mathcal{L}_{Recon}$: feature reconstruction loss, $\mathcal{R}$: regressor and $\mathcal{D_R}$: data regressor}. 
    \label{fig:model}
    \vspace{-0.3cm}
\end{figure*}

\begin{figure*}[t]
    \centering
    \includegraphics[width=0.9\textwidth]{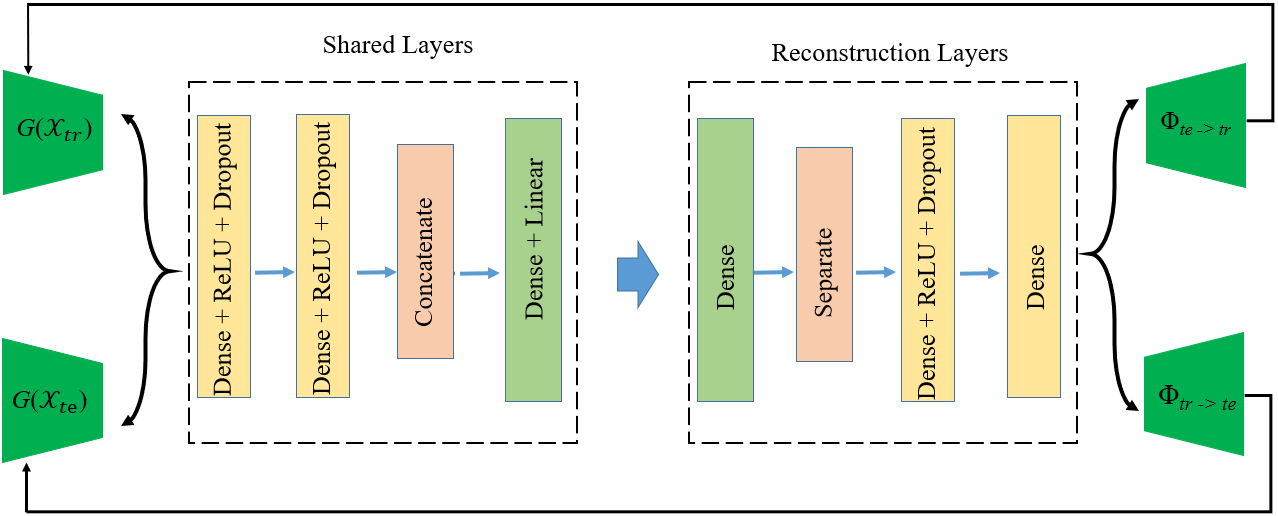}
    \caption{Feature reconstruction layers ($\mathcal{L}_{Recon}$). The shared layers are optimized by both $G(\mathcal{X}_{tr})$ and $G(\mathcal{X}_{te})$. For $G(\mathcal{X}_{tr})$, the output from reconstruction layers should be closer to $G(\mathcal{X}_{tr})$, but it is under the conversion of $\Phi_{te\shortrightarrow tr}$ since the shared layers are also optimized by $G(\mathcal{X}_{te})$, it is hence from $te$ to $tr$. Similarly, the output of reconstruction layer for $G(\mathcal{X}_{te})$ is under the conversion of  $\Phi_{tr\shortrightarrow te}$.}
    \label{fig:model_more}
    \vspace{-0.6cm}
\end{figure*}

In our framework, we employ a two-stage data discrepancy reduction procedure. We first extract bone features from any pre-trained or fine-tuned network. The distribution of training and test data are initially aligned, which reduced the discrepancy between them. We then use adversarial regression learning to find invariant features to further reduce the differences between training and test data.
$
    d_{\mathcal{H}} (\mathcal{D}_{tr}, \mathcal{D}_{te}) \approx  Dist(\mathcal{X}_{tr}, \mathcal{X}_{te}) +  Dist(G(\mathcal{X}_{tr}), G(\mathcal{X}_{te})),
$
specifically,
\begin{equation*}
\begin{aligned}
    & Dist(\mathcal{X}_{tr}, \mathcal{X}_{te}) =  ||\frac{1}{N_{tr}} \sum_{i=1}^{N_{tr}} G(\mathcal{X}_{tr}^i)- \frac{1}{N_{te}} \sum_{j=1}^{N_{te}} G (\mathcal{X}_{te}^j)||_{\mathcal{H}} \\ & 
    Dist(G(\mathcal{X}_{tr}), G(\mathcal{X}_{te})) =  ||\frac{1}{N_{tr}} \sum_{i=1}^{N_{tr}} \Phi(G(\mathcal{X}_{tr}^i))  - \frac{1}{N_{te}} \sum_{j=1}^{N_{te}} \Phi(G(\mathcal{X}_{te}^j))||_{\mathcal{H}} 
\end{aligned}
\end{equation*}
where $G$ is feature extractor from any pre-trained or fine-tuned neural network, and $\Phi$ is feature extractor from adversarial regression learning in Sec.~\ref{sec:adver}. $\mathcal{H}$ is the Reproducing Kernel Hilbert Space (RKHS) space.

\vspace{-0.3cm}
\subsection{Adversarial regression learning}\label{sec:adver}
Adversarial learning is widely used to mitigate data shift issues~\cite{ghifary2014domain,tzeng2017adversarial}. It minimizes the domain discrepancy by a feature extractor and a domain discriminator. The domain discriminator aims to distinguish the source domain from the target domain, while the feature extractor aims to learn domain-invariant representations to fool the domain discriminator. Given data representations from feature extractor $G$, we can learn a discriminator $D$, which can distinguish the two domains using the following binary cross-entropy loss function:
%
% \begin{equation}
% \begin{aligned}\label{eq:cross}
%     \mathcal{L_A}(G(\mathcal{X}_{tr}), G(\mathcal{X}_{te}))  & =  - \frac{1}{n_{tr}} \sum_{i=1}^{n_{tr}} \text{log} (1-D(G(\mathcal{X}_{tr_i}))) \\ &  - \frac{1}{n_{te}} \sum_{j=1}^{n_{te}} \text{log} (D(G(\mathcal{X}_{te_j}))) 
% \end{aligned}
% \end{equation}
\begin{equation}
\begin{aligned}\label{eq:cross}
    \mathcal{L_A}(G(\mathcal{X}_{tr}), & G(\mathcal{X}_{te}))  =  -\mathbb{E}_{x_{tr} \sim G(\mathcal{X}_{tr}) }  [\text{log} D (x_{tr})] -  \mathbb{E}_{x_{te} \sim G(\mathcal{X}_{te}) }  [\text{log} (1-D (x_{te}))]
\end{aligned}
\end{equation}
However, adversarial learning is typically applied in classification problems. The binary cross-entropy in Eq.~\ref{eq:cross} is also frequently used in improving the accuracy of classification problems. The cross-entropy loss is not proper to indicate the data difference in a regression problem. Therefore, we need a regression loss function to distinguish the training and test datasets, and we adopt the adversarial learning for the regression problem as follows. 
% Let $\epsilon_{tr}$ be the training error, $Acc_{\mathcal{L_A}}$ is the recognition accuracy of training and test data using adversarial learning with binary cross-entropy. Notice that, we aim to minimize the $\epsilon_{tr}$, while $Acc_{\mathcal{L_A}}$ pursuit a higher recognition accuracy. The overall performance $\epsilon_{tr}+ Acc_{\mathcal{L_A}} \geq \epsilon_{tr}$ will hence reduced if we use the binary cross-entropy. 
%
\begin{equation}
\begin{aligned}\label{eq:regre}
        &\mathcal{L_{AR}} (\Phi_{tr\shortrightarrow te},G(\mathcal{X}_{tr}), G(\mathcal{X}_{te}))   = \mathbb{E}_{x \sim (\mathcal{X}_{tr} \cup \mathcal{X}_{te}) } [ \ell ( \mathcal{D_R} (x),  \mathcal{Y}_\mathcal{DL})]   \\& = \frac{1}{n_{tr} + n_{te}} \sum_{k=1}^{n_{tr}+ n_{te}}  |\frac{\mathcal{Y}^k_\mathcal{{DL}} -\mathcal{Y}^{k'}_\mathcal{{DL}}}{\mathcal{Y}^k_\mathcal{{DL}} + \epsilon}|   +  |\frac{\overline{\mathcal{\mathcal{Y}_\mathcal{{DL}}}} -\overline{\mathcal{Y}^{'}_\mathcal{DL}}}{\overline{\mathcal{Y}_\mathcal{{DL}} + \epsilon }}| 
\end{aligned}
\end{equation}
where $\mathcal{L_{AR}}$ is adversarial regression loss, $\Phi_{tr\shortrightarrow te}$ is the mapping from training to test data; $\mathcal{D_R}$ is the adapted data regressor, and  $\mathcal{X}_{tr} \cup \mathcal{X}_{te}$ contains all training and test samples; $\mathcal{Y}_\mathcal{{DL}}$ is the data label in Fig.~\ref{fig:model}, specifically, $0$ is the label for the training data and $1$ is the label for the test data; $\mathcal{Y}^{k'}_\mathcal{{DL}}$ is the prediction of data label;  $\overline{\mathcal{Y}_\mathcal{DL}}$ and $\overline{\mathcal{Y}^{'}_\mathcal{DL}}$ denote the mean actual and predicted values of training or test dataset and $\epsilon$ is a small number (1e-9) to prevent division by 0. $\mathcal{L_{AR}}$ can measure how well-matched the training and test data are from both individual and overall differences. A perfect regression model would have an adversarial regression loss of 0.

To this end, the data regressor  $\mathcal{D_R}$ learns the data discrepancy by maximizing adversarial regression loss $\mathcal{L_{AR}}$ with the fixed $\Phi$, and the feature extractor $\Phi$ aims to learn domain-invariant representations via minimizing $\mathcal{L_{AR}}$ with the optimal regressor  $\mathcal{R}$. Eq.~\ref{eq:regre} guarantees that $tr\shortrightarrow te$, that is, given training samples, it will learn a map from training to test samples, while minimizing the $\mathcal{L}_{tr} (\mathcal{X}_{tr}, \mathcal{Y}_{tr})$ in Eq.~\ref{eq:training}.  However, Eq.~\ref{eq:regre} only guarantees training data close to test data, and it does not ensure that $\Phi_{te\shortrightarrow tr}$ maintains the features of the training samples. We hence introduce another mapping from test data to training data $\Phi_{te\shortrightarrow tr}$ and train it with the same adversarial regression loss as in  $\Phi_{tr\shortrightarrow te}$ as shown in Eq.~\ref{eq:regre2}. The only difference is that the  $1$ is the new data label for training data, and $0$ is the new data label for test data.
\begin{equation}
\begin{aligned}\label{eq:regre2}
    \mathcal{L_{AR}} (\Phi_{te\shortrightarrow tr},G(\mathcal{X}_{tr}), G(\mathcal{X}_{te})) 
\end{aligned}
\end{equation}
Therefore, we define the adversarial regression loss as:
\begin{equation}
\begin{aligned}\label{eq:regre3}
   \mathcal{L_{AR}} (G(\mathcal{X}_{tr}), G(\mathcal{X}_{te})) & = \mathcal{L_{AR}} (\Phi_{tr\shortrightarrow te},G(\mathcal{X}_{tr}), G(\mathcal{X}_{te})) \\ & +  \mathcal{L_{AR}} (\Phi_{te\shortrightarrow tr},G(\mathcal{X}_{tr}), G(\mathcal{X}_{te})) 
\end{aligned}
\end{equation}

To encourage the training and test information to be preserved during the adversarial regression learning, we propose a feature reconstruction loss in our model. Details of the feature reconstruction layers are shown in Fig.~\ref{fig:model_more}; the reconstruction layers are right behind the shared layers, and it aims to reconstruct extracted features and maintain feature consistency during the conversion process without losing features. The feature reconstruction loss is defined as:
\begin{equation}
\begin{aligned}\label{eq:all1}
    \mathcal{L}&_{Recon}  (\Phi_{tr\shortrightarrow te}, \Phi_{te\shortrightarrow tr}, G(\mathcal{X}_{tr}), G(\mathcal{X}_{te})) \\ &= \mathbb{E}_{x_{tr} \sim G(\mathcal{X}_{tr}) }  [\ell'(\Phi_{te\shortrightarrow tr} (\Phi_{tr\shortrightarrow te} (x_{tr})) - x_{tr} )] \\ &
    + \mathbb{E}_{x_{te} \sim G(\mathcal{X}_{te}) }  [\ell'(\Phi_{tr\shortrightarrow te} (\Phi_{te\shortrightarrow tr} (x_{te})) - x_{te} )] 
\end{aligned}
\end{equation}
where $\ell'$ is the mean square error loss, which calculates the difference between true features and reconstructed features.

\vspace{-0.3cm}
\subsection{Overall objective}
We combine the three loss functions to formalize our objective function:
\begin{equation}
\begin{aligned}\label{eq:loss_all}
\mathcal{L} (\mathcal{X}_{tr}, \mathcal{X}_{te}, \mathcal{Y}_{tr}, \Phi_{tr\shortrightarrow te}, \Phi_{te\shortrightarrow tr} )  &  = \mathcal{L}_{tr} (\mathcal{X}_{tr}, \mathcal{Y}_{tr})+ \beta \mathcal{L_{AR}} (G(\mathcal{X}_{tr}), G(\mathcal{X}_{te}))   \\& + \gamma \mathcal{L}_{Recon} (\Phi_{tr\shortrightarrow te}, \Phi_{te\shortrightarrow tr}, G(\mathcal{X}_{tr}), G(\mathcal{X}_{te})),
\end{aligned}
\end{equation}
where $\beta$ and $\gamma$ are trade-off parameters. Our model ultimately solves the following optimization problem. It minimizes the difference during the transition from the training to test data and from test to training data. Meanwhile, it maximizes the identification ability of training or test data.\footnote{Source code is available at \url{https://github.com/YoushanZhang/Adversarial-Regression-Learning-for-Bone-Age-Estimation}.}  
\begin{equation*}
\begin{aligned}\label{eq:loss_o}
   \mathop{\argmin} \min_{\substack{\Phi_{tr\shortrightarrow te}\\ \Phi_{te\shortrightarrow tr}}} \max_{\mathcal{D}_{tr}, \mathcal{D}_{te}}\  & \mathcal{L} (\mathcal{X}_{tr}, \mathcal{X}_{te}, \mathcal{Y}_{tr}, \Phi_{tr\shortrightarrow te}, \Phi_{te\shortrightarrow tr} ) 
\end{aligned}
\end{equation*}

\vspace{-0.6cm}
\section{Experiments}
We primarily validate our methods using the Bone-Age dataset. To demonstrate the generalizability of the architecture of $ARLNet$, we also evaluate our approach on the task of predicting age using two face datasets.  
\vspace{-0.3cm}
\subsection{Datasets}

\vspace{-0.1cm}
\paragraph{Bone-Age} includes data from the 2017 Pediatric Bone Age Challenge, which is organized by the Radiological Society of North America (RSNA). The statistics of the Bone-Age dataset are shown in Tab.~\ref{tab:sta} (Note that the test data is clearly different from the training data.) It also includes gender information associated with the bone age.  
\vspace{-0.2cm}
\paragraph{Face-Age} is from the UTKFace dataset, which contains 9780 images with ages from 1 to 116. It also includes gender information with its associated images (more details can be found in Zhang et al.~\cite{zhang2017age}). We also consider age regression in this dataset.  The statistics of Face-Age dataset are shown in Tab.~\ref{tab:sta_face}.

%
% \vspace{-0.9cm}
\begin{table}[t]
% \small 
\begin{center}
\captionsetup{font=small}
\caption{Statistics of Bone-Age dataset}
% \vspace{-0.3cm}
\setlength{\tabcolsep}{+1.8mm}{
\begin{tabular}{rlllllll}
\hline \label{tab:sta}
Bone-Age & \# Total  & \# Male & \# Female  & Bone ages \\
\hline
Training set & 12611 & 6833  & 5778 & 10.8 $\pm$ 3.5 \\
Validation set & 1425 & 773  & 652 & 10.8 $\pm$ 3.5 \\
Test set & 200 & 100  & 100 & 8.8 $\pm$ 3.6 \\
\hline
\end{tabular}}
\end{center}
\vspace{-1cm}
\end{table}

\vspace{-0.1cm}
\begin{table}[t]
% \small 
\begin{center}
\captionsetup{font=small}
\caption{Statistics of Face-Age dataset}
% \vspace{-0.3cm}
\setlength{\tabcolsep}{+1.0mm}{
\begin{tabular}{rlllllll}
\hline \label{tab:sta_face}
Face-Age  & \# Total  & \# Male & \# Female  & Face ages \\
\hline
Training set & 4890  & 2183  & 2707 & 29.43 $\pm$ 24.79 \\
Test set & 4890 & 2189   & 2701 & 29.41 $\pm$ 24.76 \\
\hline
\end{tabular}}
\end{center}
\vspace{-0.9cm}
\end{table}

\paragraph{MORPH II} is from MORPH database, which contains more than 55,000 face images of 13,000 individuals aged from 16 to 77 years. It includes images with detailed age, gender, and many ethnicities).
%: Black, White, Asian, Hispanic, and others). 
We followed the training/testing settings of~Shen et al.~\cite{shen2018deep} in the experiments, which selects 5,492 images of Caucasians. 
%These images are randomly split into two subsets: a training set of 80\% of the images, and a testing set containing the remaining images. 
The final performance is reported using five-fold cross-validation.

\subsection{Implementation details}

In the Bone-Age dataset, the features are extracted from an Inception V3 neural network through the last fully connected layer. 
%The fine-tuned Inception V3 model ends with 
We attach
a regression layer to output the bone age, and gender is also another input for the model. One represented feature vector has the size of $1\times 1000$ and is corresponding to one bone image. For the Bone-Age dataset, the  represented feature vectors for training, validation, and test data have a size of $11611 \times 1000$, $1425  \times 1000$, and $200 \times 1000$, respectively. Similarly, we extract features from the last fully connected layer in a pre-trained Inception V3 neural network for the Face-Age dataset. The size of feature vectors of training and test data is $4890 \times 1000$.  For MORPH II, the size of feature vectors of training and test data are  $4394 \times 1000$, and $1098 \times 1000$, respectively. We then train the model based on these extracted feature vectors.  The parameters of $ARLNet$ are first tuned based on the performance of the validation dataset of the Bone-Age dataset. We then applied these parameters to test data of Bone-Age, Face-Age and MORPH II datasets.

The numbers of units of the dense layer in shared layers are 512, 8, and 1, while the numbers of units in the reconstruction layers are opposite (1, 8, and 512). The dropout rate is 0.5. Our implementation is based on Keras and 
%our network is trained on an NVIDIA Geforce 1080 Ti equipped with 11Gb of memory. The 
the parameters settings are $\alpha = 10$,  $\beta=\gamma = 0.5$, learning rate: $\eta= 0.0001$, batch size = 128, number of iterations is 300 and the optimizer is Adam. We first train the regression model using the percentage error loss on the labeled training data. Next, we perform adversarial regression learning using the extracted features and feature reconstruction loss, which yields the learned parameters for the feature transformations $\Phi_{tr\shortrightarrow te}$ and $\Phi_{te\shortrightarrow tr}$. We also compare our results with 15 state-of-the-art methods (including both traditional methods and deep neural networks). All re-implemented methods are marked in bold in Tables~\ref{tab:re} and \ref{tab:res}.  

\subsection{Evaluation}

We use mean absolute error to evaluate our model:
$
    \text{MAE}= \frac{1}{N} \sum\nolimits_{i=1}^{N} (|y_i-\hat{y_i}|),
$
where $y$ is the provided 
%bone or face 
age, and $\hat{y}$ is the predicted age. 
%Note that a smaller MAE value means less error
%%between real and predicted age, 
%and thus a better model. 

\vspace{-0.2cm}
\paragraph{Bone-Age.} The comparison of MAE is listed in Tab.~\ref{tab:re}. We find that 
%our proposed 
$ARLNet$ has the lowest MAE versus all other approaches (23\% reduction from the best baseline).
%, which reveals our model outperforms other competing baselines. 
Specifically, online software has the worst performance among all methods. It is a pruned version of Inception V3~\cite{halabi2019rsna}, which excludes  gender information, and the difference in performance between the two models is significant (more than 65\% reduction). The MAE of Inception V3 is 4.20 months, while MAE of the pruned version is 12.35 months. This demonstrates that gender is important in predicting bone age. The performance of online software could be regarded as a lower bound. SVR is a traditional method, and its performance relies on the extracted features; it also does not consider the discrepancy between training and test data.  GSM has better results than the SVR model since it samples more data between training and test data, which reduces the discrepancy between them. The performance of pre-trained models (VGG16 and Xception) also leads to higher MAE values. The is because the pre-trained model is trained based on the ImageNet dataset, while it does not contain information from bone images. In addition, we observe that two similar methods, DANN and ADDA, also have high MAE values. There are two possible reasons. First, the feature extraction layers of DANN and ADDA models are too shallow to extract detailed information for radiology  bone images. Second, the adversarial learning of these two models only considers the transition from training data to test data.   %in Eq.~\ref{eq:cross}

\begin{table}[t]
\parbox{.45\linewidth}{
\begin{center}
\captionsetup{font=small}
\caption{Results comparison of Bone-Age dataset}
\setlength{\tabcolsep}{+0.7mm}{
\begin{tabular}{rlllllll}
\hline \label{tab:re}
Methods & MAE \\
\hline
\textbf{Online software}~\cite{Cicero2017} &  12.35 \\
\textbf{SVR}~\cite{drucker1997support} &   10.70  \\
\textbf{VGG16}~\cite{simonyan2014very} & 7.57\\
\textbf{GSM}~\cite{zhang2019transductive} & 6.06 \\
\textbf{DANN}~\cite{ghifary2014domain} & 5.26\\
\textbf{ADDA}~\cite{tzeng2017adversarial} & 4.91\\
\textbf{Xception}~\cite{chollet2017xception} & 4.39\\
\hline
\hline
ResNet50~\cite{larson2018performance} & 6.00\\
U-Net-VGG~\cite{iglovikov2018paediatric} &  4.97\\
% ResNet50~\cite{halabi2019rsna} & 4.40\\
Ice Module~\cite{halabi2019rsna} & 4.40\\
Inception V3~\cite{halabi2019rsna} & 4.20\\
Ensembles Models~\cite{pan2019improving} &  3.93 \\
\hline
\hline
\textbf{$ARLNet$} & \textbf{3.01} \\
\hline
\end{tabular}}
\end{center}
}
\hspace{+0.3cm}
\parbox{.45\linewidth}{
\begin{center}
\captionsetup{font=small}
\caption{MAE comparison on Face-Age and MORPH II datasets }
\setlength{\tabcolsep}{+0.7mm}{
\begin{tabular}{rccc}
\hline \label{tab:res}
Methods & Face-Age & MORPH II \\
\hline
% \textbf{Googlenet}~\cite{szegedy2015going} &12.30 \\
\textbf{VGG16}~\cite{simonyan2014very} & 10.40 & 5.37\\
\textbf{VGG19}~\cite{simonyan2014very} & 9.64 & 4.93\\
% \textbf{Densenet201}~\cite{huang2017densely} & 9.61 \\
\textbf{SVR}~\cite{drucker1997support} &   9.10  & 5.77 \\
% \textbf{Resnet101}~\cite{he2016deep} & 8.74\\
\textbf{ResNet50}~\cite{he2016deep} & 8.64 & 4.02\\
\textbf{Xception}~\cite{chollet2017xception} & 8.49 & 3.88\\
\textbf{Inception V3}~\cite{chollet2017xception} & 8.36  & 3.65\\
\textbf{GSM}~\cite{zhang2019transductive} & 8.06 & 3.35\\
\textbf{DANN}~\cite{ghifary2014domain} & 6.26 & 3.01\\
\textbf{ADDA}~\cite{tzeng2017adversarial} & 5.91 & 2.73\\
\hline
\hline
ARN~\cite{agustsson2017anchored} & - & 3.00\\
DRFs~\cite{shen2018deep} & - & 2.91\\
DCNN~\cite{dornaika2020robust} & - & 2.75\\
\hline
\hline
\textbf{$ARLNet$} & \textbf{4.50} & \textbf{2.28} \\
\hline
\end{tabular}}
\end{center}
}
\vspace{-0.6cm}
\end{table}

% \begin{table}[t]
% % \small 
% \begin{center}
% \captionsetup{font=small}
% \caption{MAE comparison on Face-Age and MORPH II datasets }
% \setlength{\tabcolsep}{+3.7mm}{
% \begin{tabular}{rccc}
% \hline \label{tab:res}
% Methods & Face-Age & MORPH II \\
% \hline
% % \textbf{Googlenet}~\cite{szegedy2015going} &12.30 \\
% \textbf{VGG16}~\cite{simonyan2014very} & 10.40 & 5.37\\
% \textbf{VGG19}~\cite{simonyan2014very} & 9.64 & 4.93\\
% % \textbf{Densenet201}~\cite{huang2017densely} & 9.61 \\
% \textbf{SVR}~\cite{drucker1997support} &   9.10  & 5.77 \\
% % \textbf{Resnet101}~\cite{he2016deep} & 8.74\\
% \textbf{ResNet50}~\cite{he2016deep} & 8.64 & 4.02\\
% \textbf{Xception}~\cite{chollet2017xception} & 8.49 & 3.88\\
% \textbf{Inception V3}~\cite{chollet2017xception} & 8.36  & 3.65\\
% \textbf{GSM}~\cite{zhang2019transductive} & 8.06 & 3.35\\
% \textbf{DANN}~\cite{ghifary2014domain} & 6.26 & 3.01\\
% \textbf{ADDA}~\cite{tzeng2017adversarial} & 5.91 & 2.73\\
% \hline
% \hline
% ARN~\cite{agustsson2017anchored} & - & 3.00\\
% DRFs~\cite{shen2018deep} & - & 2.91\\
% DCNN~\cite{dornaika2020robust} & - & 2.75\\
% \hline
% \hline
% \textbf{$ARLNet$} & \textbf{4.50} & \textbf{2.28} \\
% \hline
% \end{tabular}}
% \end{center}
% \vspace{-0.6cm}
% \end{table}
%

\vspace{-0.3cm}
\paragraph{Face-Age.}\hspace{-.02in}The comparison of MAE is listed in Tab.~\ref{tab:res}. $ARLNet$ again has the lowest MAE value, outperforming all others (24\% reduction than the best baseline). We notice that results from Inception V3 are close to our model since our model is based on extracted features from the pre-trained Inception V3. However, our model has lower MAE than Inception V3 model, which demonstrates that the adversarial regression learning is useful in the regression problem. We find that the SVR model is better than some neural networks (VGG16 and Googlenet) since SVR uses the features from the Inception V3 model. SVR normally has a higher error than the fine-tuned IncetionV3 network. In addition, the performance of two domain adaptation methods (DANN and ADDA) have results close to that of our model. The underlying reason is that face images are easier to find data discrepancy and extract features since images are RGB images, while radiology bone images are significantly different from RGB images, it is difficult for these two methods to extract fine-grained bone features. 

\vspace{-0.3cm}
\paragraph{MORPH II.}\hspace{-.02in}Tab.~\ref{tab:res} also compares performance between $ARLNet$ and  state-of-the-art methods. 
%We find that o
ARLNet achieves the best performance, reducing the error rate by 17.1\% compared to the best baseline model (DCNN), suggesting that adversarial regression learning is effective in reducing the discrepancy between training and test data, and achieves lower MAE. 
%One interesting observation is that the simple domain adaptation method ADDA is slightly better than the DCNN model, which illustrates that data discrepancy is large in Caucasian population.    

% \begin{figure*}[h!]
% \centering
% \begin{subfigure}{0.25\textwidth}
% \includegraphics[width=\linewidth]{Figures/valid1.png}
% \caption{$G(\mathcal{X}_{va})$} \label{fig:ima}
% \end{subfigure}
% \begin{subfigure}{0.23\textwidth}
% \includegraphics[width=\linewidth]{Figures/valid2.png}
% \caption{$\Phi(G(\mathcal{X}_{va}))$} \label{fig:imb}
% \end{subfigure} 
% \begin{subfigure}{0.25\textwidth}
% \includegraphics[width=\linewidth]{Figures/test1.png}
% \caption{$G(\mathcal{X}_{te})$} \label{fig:imc}
% \end{subfigure}
% \begin{subfigure}{0.23\textwidth}
% \includegraphics[width=\linewidth]{Figures/test22.png}
% \caption{$\Phi(G(\mathcal{X}_{te}))$} \label{fig:imd}
% \end{subfigure}
% \caption{The t-SNE visualizations of features generated by $ARLNet$ in Bone-Age dataset.  Fig.~\ref{fig:ima} and~\ref{fig:imc} are extracted features from fine-tuned Inception V3 before training. Fig.~\ref{fig:imb} and~\ref{fig:imd} are extracted features using $ARLNet$ after training. Different ages are significantly aligned after the adversarial regression learning to reduce the data discrepancy.   Colorbar means different bone ages in the dataset ($\mathcal{X}_{va}$ is the validation data).} \label{fig:sne}
% \vspace{-0.3cm}
% \end{figure*}

% \vspace{-0.3cm}
\section{Discussion}

There are three 
%compelling 
reasons that our model outperforms state-of-the-art methods. First, the proposed percentage loss $\ell$ is able to consider both the individual and mean percentage error. Second, we propose adversarial regression loss, which can maintain transition from training data to test data and transition from test data to training data. Third, we propose feature reconstruction loss, which further guarantees the consistency of training and test samples.   
% However, one weakness of our model is that we evaluate our model only on two large-scale datasets: Bone-Age and Face-Age, test on other regression data will improve the applicability of our model. 

\begin{table}[b]
\small
\begin{center}
\vspace{-0.3cm}
\captionsetup{font=small}
\caption{Ablation experiments of effects of different loss functions of Bone-Age dataset}
\setlength{\tabcolsep}{+6.4mm}{
\begin{tabular}{rlllllll}
\hline \label{tab:Abla}
Methods & MAE \\
\hline
$ARLNet$$-\mathcal{L}_{D}-\mathcal{L}_{AR}-\mathcal{L}_{Recon}$ & 4.52\\
$ARLNet$$-\mathcal{L}_{M}-\mathcal{L}_{AR}-\mathcal{L}_{Recon}$ &4.38\\
$ARLNet$$-\mathcal{L}_{D}-\mathcal{L}_{Recon}$ & 4.17\\
$ARLNet$$-\mathcal{L}_{M} -\mathcal{L}_{Recon}$ & 4.03\\
$ARLNet-\mathcal{L}_{AR} - \mathcal{L}_{Recon}$ & 3.91\\
$ARLNet-\mathcal{L}_{AR}$ & 3.45\\
$ARLNet-\mathcal{L}_{Recon}$ & 3.26\\
\hline
\hline
$ARLNet$ & \textbf{3.01} \\
\hline
\end{tabular}}
\end{center}
\vspace{-0.8cm}
\end{table}

\vspace{-0.3cm}
\paragraph{Ablation study.}\hspace{-.1in}
To better demonstrate the performance of our model, we report the effects of different loss functions on classification accuracy in Tab.~\ref{tab:Abla} ($\mathcal{L_D}$:  absolute mean discrepancy loss, $\mathcal{L_M}$: mean absolute percentage error loss, and $\mathcal{L_{AR}}$: adversarial regression loss and $\mathcal{L}_{Recon}$: feature reconstruction loss). ``$ARLNet-\mathcal{L}_{D}-\mathcal{L}_{AR}-\mathcal{L}_{Recon}$” is implemented without absolute mean discrepancy loss, adversarial regression loss, and feature reconstruction loss. It is a simple model, which only considers the labeled training data using mean absolute percentage error loss. ``$ARLNet-\mathcal{L}_{AR} - \mathcal{L}_{Recon}$” reports results without performing the additional adversarial regression loss and feature reconstruction loss. ``$ARLNet-\mathcal{L}_{Recon}$” trains the model with percentage error loss and adversarial regression loss without the feature reconstruction loss. We observe that with the increasing of the number of loss functions, the robustness of our model keeps improving.  Therefore, we can conclude that all these different loss functions are important in maximizing regression performance.
%minimizing the discrepancy between training and test data.

% To show the progress of our $ARLNet$, we also visualize data before training and after training as shown in Fig.~\ref{fig:sne}. We notice that the bone ages are mixing with each other before training in Fig.~\ref{fig:ima} and~\ref{fig:imc}, which denotes that $G(\mathcal{X}_{va})$ and  $G(\mathcal{X}_{te})$ are difficult to directly estimate bone age. However, the generated features from $ARLNet$ is significantly better aligned in Fig.~\ref{fig:imb} and~\ref{fig:imd} after training, and we clearly observe the pattern of bone ages changes from small value to a large value.  Therefore, $ARLNet$ is effective in reducing the data discrepancy.  

\vspace{-0.2cm}
\paragraph{What can we learn from {\em ARLNet}?} As shown in the ablation study, we know the effects of different loss functions. The adversarial regression loss has a dominant effect on the final results. Differing from traditional machine learning that only optimizes models using training data, our $ARLNet$ considers transductive learning~\cite{zhang2019transductive}, and reduces the discrepancy between training and test data. It will be useful to include the test data during the training without any labels. Although there are no labels for the test data, the discrepancy between training samples and test samples is minimized, the test risk is thus reduced if there is a difference between training and test data. Therefore, the performance of the model will be improved if we feed the test data during the training processes with ARL.     

% \vspace{-0.3cm}
\section{Conclusion}

We have presented an adversarial regression learning network ($ARLNet$) for bone age estimation that reduces the discrepancy between training and test data.  In particular, we consider the problem from the traditional training protocol to adversarial regression learning.  The adversarial regression learning consists of adversarial regression and feature reconstruction losses. The adversarial regression loss can push the prototype of bone ages computed in either training or test data close in the embedding space,  and maintain invariant representations across two datasets. In addition, the proposed feature reconstruction loss further guarantees the structure and content from training and test data, and it will take the decision of regressor into account to align feature distribution, which leads to domain-invariant representations.
Our approach provides a 
%consistent 
more than 20\% error reduction over the state of the art in 
%distinct 
two
age regression tasks.

\bibliographystyle{unsrt}
\bibliography{references}
\end{document}